\documentclass[11pt,a4paper]{article}
\usepackage[hyperref]{acl2017}
\usepackage{amsmath, amsthm, amssymb}
\usepackage{times}

\usepackage{url}

\usepackage{multirow}
\usepackage{booktabs}
\usepackage{graphicx}
\usepackage{subfigure}

\usepackage{pgfplots}
\pgfplotsset{width=0.48\textwidth,compat=1.9, height=2.3in}

\usepackage[normalem]{ulem}

\aclfinalcopy % Uncomment this line for the final submission
\newcommand{\algname}{\textit{arc-swift}}
\newcommand{\arceager}{\textit{arc-eager}}
\newcommand{\arcstandard}{\textit{arc-standard}}
\newcommand{\archybrid}{\textit{arc-hybrid}}
\newcommand{\Algname}{\textit{Arc-swift}}
\newcommand{\Arceager}{\textit{Arc-eager}}
\newcommand{\Arcstandard}{\textit{Arc-standard}}
\newcommand{\Archybrid}{\textit{Arc-hybrid}}

\newcommand{\diffadd}[1]{#1}
\newcommand{\diffsub}[1]{}
\newcommand{\gitrepo}{\url{https://github.com/qipeng/arc-swift}}

\title{\emph{Arc-swift:} A Novel Transition System for Dependency Parsing}

\author{Peng Qi \qquad   Christopher D. Manning\\
  Computer Science Department \\
  Stanford University \\
  {\tt \{pengqi, manning\}@cs.stanford.edu} \\}

\date{}

\begin{document}
\maketitle

% abstract used to take to line 35
\begin{abstract}
Transition-based dependency parsers often need sequences of local shift and reduce operations to produce certain attachments.
% and hence to derive the correct parse tree for a sentence. 
Correct individual decisions hence require \diffsub{much }global information about the sentence context and mistakes cause error propagation.
  %This limits their ability to fully leverage context information available in the parser states to make transition directly leading to attachments, which impacts parsing performance in greedy parsing. 
This paper proposes a novel transition system, \algname, that enables direct attachments between tokens farther apart with a single transition. This allows the parser to leverage lexical information more directly in transition decisions. %Specifically, instead of considering a limited context of only a few tokens on the stack, our transition system allows direct attachment of a token in the buffer with any token on the stack that satisfy certain preconditions. 
  %By deferring ambiguity resolution to the parser, parsers using \algname\ achieve significantly better performance compared to those using traditional transition systems with a very small beam size. 
 % While traditional transition systems require a larger beam size for ambiguity resolution, 
Hence, \algname\ can achieve significantly better performance with a very small beam size.
% by deferring this to the parsers.
  Our parsers reduce error by 3.7--7.6\% relative to those using existing transition systems on the Penn Treebank dependency parsing task and English Universal Dependencies.
\end{abstract}

\section{Introduction}

Dependency parsing is a longstanding natural language processing task, with its outputs crucial to various downstream tasks including relation extraction \citep{schmitz2012open, angeli2015leveraging}, language modeling \citep{gubbins2013dependency}, and natural logic inference \citep{bowman2016fast}. 

Attractive for their linear time complexity and amenability to conventional classification methods, \emph{transition-based} dependency parsers \diffsub{attracted }\diffadd{have sparked} much research interest recently. %A transition-based parser models parsing as a sequential process. 
% (shortening) Inspired by shift-reduce parsers for context-free grammars \cite{aho1986compilers}, 
A transition-based parser makes sequential predictions of transitions between states under the restrictions of a \emph{transition system} \cite{nivre2003anefficient}. 
%Operating at linear time in the number of tokens in the sentence, t
Transition-based parsers have been shown to excel at parsing shorter-range dependency structures, as well as languages where non-projective parses are less pervasive \cite{mcdonald2007characterizing}. 

\begin{figure}
\centering
\includegraphics[width=.5\textwidth]{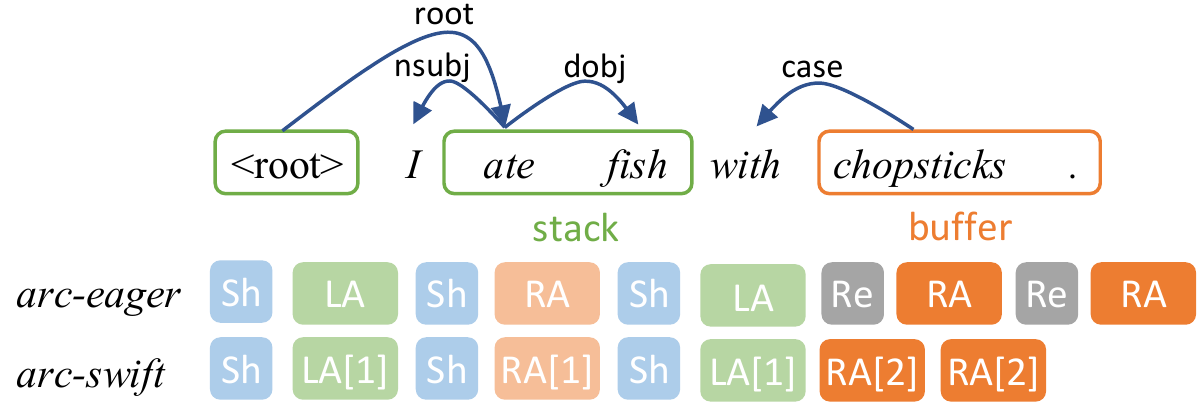}
\caption{An example of the state of a transition-based dependency parser, and the transition sequences used by \arceager\ and \algname\ 
% , respectively, 
to induce the correct parse. The state shown is generated by the first six transitions of both systems.}\label{fig:example}
\end{figure}

However, the transition systems employed in state-of-the-art dependency parsers usually define very local transitions. At each step, only one or two words are affected, with very local %changes in the parser state. 
attachments made.
As a result, distant attachments require long and not immediately obvious transition sequences %to establish 
(e.g., \textit{ate}$\to$\textit{chopsticks} in Figure~\ref{fig:example}, which requires two transitions). This is further aggravated by the usually local lexical information leveraged to make transition predictions \cite{chen2014fast, andor2016globally}. %relatively long and not immediately obvious sequences of transitions are made to establish attachments between tokens, and this confusion is further aggravated by the usually local lexical information leveraged to make transition predictions, as we will discuss in detail in Sections \ref{sec:transparse} and \ref{sec:arcswift}. 

In this paper, we introduce a novel transition system, \algname, which defines non-local transitions that directly induce attachments of distance up to $n$ ($n$ = the number of tokens in the sentence). %Our transition system can be viewed as an extension of the {\it arc-eager} system \cite{nivre2003anefficient, nivre2008algorithms}, but has the advantage of allowing direct attachments between the word on the buffer and any eligible word on the stack, which resolves potential ambiguity within a single transition instead of a sequence of transitions.
Such an approach is connected to \emph{graph-based} dependency parsing, in that it leverages pairwise scores between tokens in making parsing decisions \cite{mcdonald2005nonprojective}.

\begin{figure*}
	\centering
	\begin{tabular}{@{}ll@{}}
		\toprule
		\arcstandard & \archybrid \\
		$\begin{array}{@{}ll@{}}
		\textbf{Shift} & (\sigma, i|\beta, A) \Rightarrow (\sigma|i, \beta, A) \\
		\textbf{LArc} & (\sigma|i|j, \beta, A) \Rightarrow (\sigma|j, \beta, A \cup \{(j\to i)\}) \\
		\textbf{RArc} & (\sigma|i|j, \beta, A) \Rightarrow (\sigma|i, \beta, A\cup\{(i\to j)\})
		\end{array}$ & 
		$\begin{array}{@{}ll@{}}
		\textbf{Shift} & (\sigma, i|\beta, A)  \Rightarrow  (\sigma|i, \beta, A) \\
		\textbf{LArc} & (\sigma|i, j|\beta, A)  \Rightarrow  (\sigma, j|\beta, A \cup \{(j\to i)\}) \\
		\textbf{RArc} & (\sigma|i|j, \beta, A)  \Rightarrow  (\sigma|i, \beta, A\cup\{(i\to j)\})
		\end{array}
		$
		\\
		\midrule
		\arceager & \algname \\
		$\begin{array}{@{}ll@{}}
		\textbf{Shift} & (\sigma, i|\beta, A)  \Rightarrow  (\sigma|i, \beta, A) \\
		\textbf{LArc} & (\sigma|i, j|\beta, A)  \Rightarrow (\sigma, j|\beta, A \cup \{(j\to i)\}) \\
		\textbf{RArc} & (\sigma|i, j|\beta, A)  \Rightarrow (\sigma|i|j, \beta, A\cup\{(i\to j)\}) \\
		\textbf{Reduce} & (\sigma|i, \beta, A)  \Rightarrow  (\sigma, \beta, A)
		\end{array}$ & $\begin{array}{@{}ll@{}}
		\textbf{Shift} & (\sigma, i|\beta, A)  \Rightarrow  (\sigma|i, \beta, A) \\
		\textbf{LArc}[k] & (\sigma|i_k|\ldots|i_1, j|\beta, A) \\
		&\qquad \Rightarrow  (\sigma, j|\beta, A \cup \{(j\to i_k)\}) \\
		\textbf{RArc}[k] & (\sigma|i_k|\ldots|i_1, j|\beta, A) \\ 
		& \qquad \Rightarrow (\sigma|i_k|j, \beta, A\cup\{(i_k\to j)\})
		\end{array}
		$\\
		\bottomrule
	\end{tabular}
	\caption{Transitions defined by different transition systems.} \label{fig:transsys}
\end{figure*}

We make two main contributions in this paper. Firstly, we introduce a novel transition system for dependency parsing, which alleviates the difficulty of distant attachments in previous systems by allowing direct attachments anywhere in the stack. Secondly, we compare parsers by the number of mistakes they make in common linguistic \diffsub{categories}\diffadd{constructions}. We show that \algname\ parsers  reduce errors in attaching prepositional phrases and conjunctions compared to parsers using existing transition systems.
% To the best of our knowledge, this is the first time such an analysis has been done to compare dependency parsers.
%Lastly, we also show a side-by-side comparison of three popular traditional transition systems ({\it arc-standard}, {\it arc-eager}, and {\it arc-hybrid}) with the proposed system, in hope to provide practical guidance for future work on dependency parsing.

%This paper is organized as follows. In Section \ref{sec:transparse}, we will briefly review transition-based dependency parsing as well as three popular transition systems widely used in the literature, before we introduce our novel transition system in Section \ref{sec:arcswift}. Experiments and analyses on Penn Treebank and universal dependencies is presented in Section \ref{sec:experiments}. Finally, we review related work in Section \ref{sec:related_work}, before we conclude in Section \ref{sec:conclusion}.

%Graph-based dependency parsers attempt to assign scores to each pair of tokens in a given sentence, and outputs the final parse tree by maximizing the sum of these pair-wise scores over all possible candidate parses. Despite the simplicity of this assumption, graph-based parsers have achieved stellar performance in many dependency tasks, especially in languages where parse trees are often non-projective. When projectivity of the parse is required, however, finding a projective parse is often costly \citethis{Eisner algo}, or requires extra heuristics \cite{mcdonald2005nonprojective}.

\section{Transition-based Dependency Parsing} \label{sec:transparse}

Transition-based dependency parsing is performed by predicting transitions between states (see Figure \ref{fig:example} for an example). Parser states are usually written as $(\sigma|i, j|\beta, A)$, where $\sigma|i$ denotes the \emph{stack} with token $i$ on the top, $j|\beta$ denotes the \emph{buffer} with token $j$ at its leftmost, and $A$ the set of dependency arcs. 
%Parsing starts with $([\langle\text{root}\rangle], S, \emptyset)$ and stops when all tokens except $\langle$root$\rangle$ has been attached, where $\langle$root$\rangle$ is an artificial token signifying the root of the tree, and $S$ is all of the tokens in the sentence.
%Transition-based dependency parsing models parsing as a sequence of transitions between parser states. A parser state consists of a \emph{stack} that contains active tokens eligible for dependency arcs to originate from or end at, a \emph{buffer} that contains the list of words yet to be processed by the parser, and the \emph{set of arcs} that have already been established. Typically a parser state is written as $(\sigma|i, j|\beta, A)$, where $\sigma|i$ denotes the stack with token $i$ being its topmost/rightmost element, $j|\beta$ denotes the buffer with token $j$ being its leftmost element, and $A$ the set of arcs. For algorithmic convenience, an artificial $\langle$root$\rangle$ token is also introduced to mark the root of the tree. Parsing starts with the initial state $([\langle$root$\rangle], \beta_0, \emptyset)$ where $\beta_0$ contains all of the tokens in the sentence, and ends when every token (except $\langle$root$\rangle$) has been properly attached to some other token.
Given a state, the goal of a dependency parser is to predict a \emph{transition} to a new state that would lead to the correct parse. %A transition is usually defined by a pair of states, usually involving movement of tokens from the buffer to the stack, and/or introduction of an arc to the parse. 
A \emph{transition system} defines a set of transitions that are sound and complete for parsers, that is, every transition sequence would derive a well-formed parse tree, and every possible parse tree can also be derived from some transition sequence.% starting at the initial state
\footnote{We only focus on projective parses for the scope of this paper.}
% Perhaps cite Kuebler et al. dep parsing book for these properties.

\Arcstandard\ \citep{nivre2004incrementality} is one of the first transition systems proposed for dependency parsing. It defines three transitions: shift, left arc (LArc), and right arc (RArc) (see Figure \ref{fig:transsys} for definitions, same for the following transition systems), where all arc-inducing transitions operate on the stack. This system builds the parse bottom-up, i.e., a constituent is only attached to its head after it has received all of its dependents. A potential drawback is that during parsing, it is difficult to predict if a constituent has consumed all of its right dependents. \Arceager\ \cite{nivre2003anefficient} %Partly motivated by psycholinguistic concerns, \citet{nivre2003anefficient} presented \arceager, which 
remedies this drawback by defining arc-inducing transitions that operate between the stack and the buffer. As a result, a constituent no longer needs to be complete before it can be attached to its head to the left, as a right arc doesn't prevent the attached dependent from taking further dependents of its own.\footnote{A side-effect of \arceager\ is that there is sometimes \diffsub{an ambiguity}\diffadd{\emph{spurious ambiguity}} between Shift and Reduce transitions. For the example in Figure \ref{fig:example}, the first Reduce can be inserted before the third Shift without changing the correctness of the resulting parse, i.e., both are feasible at that time.} %This comes at a minor cost: certain preconditions would now need to be met before each transition is feasible to preserve soundness and completeness. LArc is only feasible if the top of the stack isn't already attached to any token in the sentence, and Reduce is only feasible if the top of the stack is already attached to some token. Formally,
%\\[.3em]$
%\begin{array}{ll}
%\textbf{LArc} & (k \to i) \notin A, \forall k\in\{0, 1,\ldots,n\} \\
%\textbf{Reduce} & \exists k\in\{0, 1, \ldots, n\}\text{ s.t. } (k \to i) \in A
%\end{array}$
%\\[.3em]
%here $n$ denotes the total number of words, and 0 stands for the $\langle$root$\rangle$ token.
\citet{kuhlmann2011dynamic} propose a hybrid system derived from a tabular parsing scheme, which they have shown both \arcstandard\ and \arceager\ can be derived from. \Archybrid\ combines LArc from \arceager\ and RArc from \arcstandard\ to build dependencies bottom-up.

\section{Non-local Transitions with \algname} \label{sec:arcswift}

T\diffadd{he t}raditional transition systems discussed in Section \ref{sec:transparse} only allow very local transitions affecting one or two words, which makes long-distance dependencies difficult to predict. %Their inherent ambiguity could also adversely affect the parser's performance if not correctly addressed. 
%Motivated by this observation, %and inspired by graph-based parsers, 
%we introduce the \algname\ transition system to address this issue.
To illustrate the limitation of local transitions, consider parsing the following sentences:

% \begin{quote}
\noindent
\hspace*{1cm}{\it I ate fish with ketchup.} \\
\hspace*{1cm}{\it I ate fish with chopsticks.}
% \end{quote}

\noindent
The two sentences have almost identical structures, % in terms of number of tokens, and the parts-of-speech thereof. 
with the notable difference that the prepositional phrase is complementing the direct object in the first case, and the main verb in the second.
%in the first sentence, the prepositional phrase ``\textit{with ketchup}'' modifies the direct object ``\textit{fish}'', where in the second ``\textit{with chopsticks}'' modifies the main verb ``\textit{ate}'' as an instrument.

\begin{figure}
	\centering
	%\subfigure[]{ \includegraphics[width=0.235\textwidth]{images/example-arc-standard} \label{fig:decision:a}
	%}\subfigure[]{ \includegraphics[width=0.235\textwidth]{images/example-arc-eager} 
	%	\label{fig:decision:b}
	%}
	\includegraphics[width=0.35\textwidth]{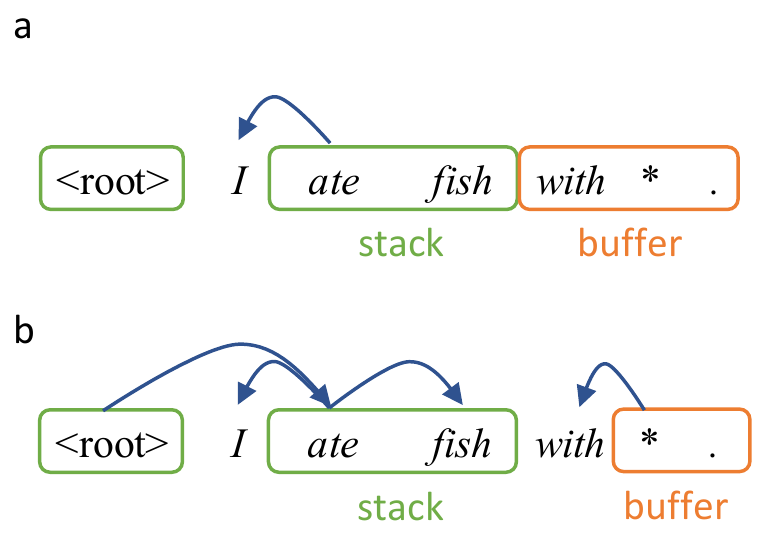}
	\caption{Parser states that present difficult transition decisions in traditional systems. In these states, parsers would need to incorporate context about ``ate'', ``fish'', and ``$\star$'' to make the correct local transition.\label{fig:decision}} 
\end{figure}

For \arcstandard\ and \archybrid, the parser would have to decide between Shift and RArc when the parser state is as shown in Figure \ref{fig:decision}a, where $\star$ stands for either ``\textit{ketchup}'' or ``\textit{chopsticks}''.\footnote{For this example, we assume that the sentence is being parsed into Universal Dependencies.} Similarly, an \arceager\ parser would deal with the state shown in Figure \ref{fig:decision}b. Making the correct transition requires information about %the context, namely 
context words ``\textit{ate}'' and ``\textit{fish}'', as well as ``$\star$''. 

Parsers employing traditional transition systems would usually incorporate more features about the context in the transition decision, or employ beam search during parsing \citep{chen2014fast, andor2016globally}. %For instance information about the top-$N$ (where $N$ is usually 3) elements in the stack, as well as various features about the working parse tree have been applied to this disambiguation .

In contrast, inspired by graph-based parsers, %which make predictions by scoring pairs of tokens for attachment, 
we propose \algname, which defines non-local transitions as shown in Figure \ref{fig:transsys}.
This allows direct comparison of different attachment points, and provides a direct solution to parsing the two example sentences. When the \algname\ parser encounters a state identical to Figure \ref{fig:decision}b, it could directly compare transitions RArc[1] and RArc[2] instead of evaluating between local transitions. This results in a direct attachment %to transition to the desired state 
much like that in a graph-based parser, informed by lexical information about affinity of the pairs of words.

\Algname\ also bears much resemblance to \arceager. In fact, an LArc$[k]$ transition can be viewed as $k-1$ Reduce operations followed by one LArc in \arceager, and similarly for RArc$[k]$. 
%Naturally, the precondition is similar to the same sequence of transitions of {\it arc-eager}, namely for an LArc$[n]$ or RArc$[n]$ transition to be feasible, we necessarily need
%\[\forall 1 \le k < n, \exists j \text{ s.t. }\{(j \to i_k)\},\]
%i.e. all of the tokens above the $n$-th token in the stack should have already been attached to some token, with the additional requirement that $i_n$ is not attached to any word for LArc$[n]$. 
Reduce is no longer needed in \algname\ as it becomes part of LArc$[k]$ and RArc$[k]$, removing the ambiguity in derived transitions in \arceager. %This also implies that at each time only one LArc$[n]$ transition is feasible for some $n$. Since all Reduce transitions in {\it arc-eager} are necessarily followed by LArc or RArc,
\algname\ is also equivalent to \arceager\ in terms of soundness and completeness.\footnote{This is easy to show because in \arceager, all Reduce transitions can be viewed as preparing for a later LArc or RArc transition. \diffadd{We also note that similar to \arceager\ transitions, \algname\ transitions must also satisfy certain pre-conditions. Specifically, an RArc$[k]$ transition requires that the top $k-1$ elements in the stack are already attached; LArc$[k]$ additionally requires that the $k$-th element is unattached, resulting in no more than one feasible LArc candidate for any parser state.}} 
% \todo{Do we need to write a short proof for this in supplementary material?}
A caveat is that the worst-case time complexity of \algname\ is $O(n^2)$ instead of $O(n)$, which existing transition-based parsers enjoy. However, in practice the runtime is nearly linear, thanks to the usually \diffsub{short stack length}\diffadd{small number of reducible tokens in the stack}.

\section{Experiments} \label{sec:experiments}

\subsection{Data and Model}

We use the Wall Street Journal portion of Penn Treebank with standard parsing splits (PTB-SD), along with Universal Dependencies v1.3 \cite{nivre2016universal} (EN-UD). PTB-SD is converted to Stanford Dependencies \cite{demarneffe2008stanford} with CoreNLP 3.3.0 \cite{manning-EtAl:2014:P14-5} following previous work. We report labelled and unlabelled attachment scores (LAS/UAS), removing punctuation from all evaluations.

%\begin{table*}
%	\centering

%	\begin{tabular}{lllllll}
%		\toprule

%		\multirow{2}{*}{Model} & \multirow{2}{*}{Transition system} &  \multirow{2}{*}{Notes} & \multicolumn{2}{c}{PTB-SD} & \multicolumn{2}{c}{EN-UD} \\
%		& & & UAS & LAS & UAS & LAS \\
%		\midrule
%		This work & \arcstandard && 93.81 & 91.57 &\\
%		This work & \arceager{\it -S} && 93.93 & 91.75 &\\
%		This work & \arceager{\it -R} && 93.95 & 91.76 &\\
%		This work & \archybrid && 93.98 & 91.74 &\\
%		This work & \algname &  & 94.23 & 92.23\\
%		\midrule
%		Andor\citeyear{andor2016globally} & \arcstandard & CRF-like loss, B=32 & \textbf{94.61} & \textbf{92.79} & 84.79* & 80.38* \\
		%		Andor\citeyear{andor2016globally} & {\it arc-standard} & B=32 & 93.59 & 91.70 \\
%		K\&G\citeyear{kiperwasser2016simple} & \archybrid & dyn-oracle, 11 BiLSTM vectors & 93.9 & 91.9 \\
		%		Weiss\citeyear{weiss2015structured} & {\it arc-standard} & & 93.19 & 91.18 \\
%		Weiss\citeyear{weiss2015structured} & \arcstandard & B=8 & 93.99 & 92.05 \\
		%		\midrule
		%		K\&G\citeyear{kiperwasser2016simple} & - & graph &  93.0 & 90.9 \\
%		\bottomrule
%	\end{tabular}
%	\caption{Performance of parsers using different transition systems on the Penn Treebank dataset. *: We obtained these from their published results online.} \label{tab:ptb}
%\end{table*}

Our model is very similar to that of \cite{kiperwasser2016simple}, where features are extracted from tokens with bidirectional LSTMs, and concatenated for classification. 
For the three traditional transition systems, features of the top 3 tokens on the stack and the leftmost token in the buffer are concatenated as classifier input.
For \algname, features of the head and dependent tokens for each arc-inducing transition are concatenated to compute scores for classification, and features of the leftmost buffer token is used for Shift.
%Specifically, for each word, we extract a ``head'' and a ``dependent'' representation. For the three traditional systems, two representations are merged by a function $f$, and the merged representations of the top 3 tokens on the stack and the leftmost token in the buffer are used for classification. For \algname, the head and dependent representation of the respective tokens are merged by $f$ for each possible transition to make classification decisions. 
For other details we defer to Appendix \ref{sec:model_training}. \diffadd{The full specification of the model can also be found in our released code online at \gitrepo.}

\subsection{Results}

%We evaluated parsers using all four transition systems on the two datasets. %We also implemented the dynamic oracle with exploration \cite{goldberg2012dynamic} for each transition system to examine how much each transition system benefits from it. For the static oracle counterpart, 
We use static oracles for all transition systems, and for \arceager\ we implement oracles that always Shift/Reduce when ambiguity is present (\arceager{\it-S/R}). We %train our parsers with static oracles and 
evaluate our parsers with greedy parsing (i.e., beam size 1).
% and one that always Reduces ({\it arc-eager-R}).
The results are shown in  Table~\ref{tab:ptb}.\footnote{In the interest of space, we abbreviate all transition systems (TS) as follows in tables: {\it asw} for \algname, {\it asd} for \arcstandard, {\it aeS/R} for \arceager{\it -S/R}, and {\it ah} for \archybrid.} Note that K\&G \citeyear{kiperwasser2016simple} is trained with a dynamic oracle \cite{goldberg2012dynamic}, Andor \citeyear{andor2016globally} with a CRF-like loss, and both Andor \citeyear{andor2016globally} and Weiss \citeyear{weiss2015structured} employed beam search (with sizes 32 and 8, respectively).

\addtocounter{footnote}{-1}

\begin{table}[t]
	\centering
	\begin{tabular}{@{}llllll@{}}
		\toprule
		\multirow{2}{*}{Model} & \multirow{2}{*}{TS} & \multicolumn{2}{c}{PTB-SD} & \multicolumn{2}{c}{EN-UD} \\
		& & UAS & LAS & UAS & LAS \\
		\midrule
		This work & {\it asd} & 94.0 & 91.7 & 85.6 & 81.5 \\
		This work & {\it aeS} & 94.0 & 91.8 & 85.4 & 81.4 \\
		This work & {\it aeR} & 93.8 & 91.7 & 85.2 & 81.2 \\
		This work & {\it ah} & 93.9 & 91.7 & 85.4 & 81.3 \\
		This work & {\it asw} & 94.3 & 92.2 & {\bf 86.1} & {\bf 82.2} \\
		\midrule
		Andor \citeyear{andor2016globally} & {\it asd} & \textbf{94.6} & \textbf{92.8} & 84.8* & 80.4* \\
		%		Andor\citeyear{andor2016globally} & {\it arc-standard} & B=32 & 93.59 & 91.70 \\
		K\&G \citeyear{kiperwasser2016simple} & {\it ah} & 93.9 & 91.9 & & \\
		%		Weiss\citeyear{weiss2015structured} & {\it arc-standard} & & 93.19 & 91.18 \\
		Weiss \citeyear{weiss2015structured} & {\it asd} & 94.0 & 92.1 & &  \\
		%		\midrule
		%		K\&G\citeyear{kiperwasser2016simple} & - & graph &  93.0 & 90.9 \\
		C\&M \citeyear{chen2014fast} & {\it asd} & 91.8 & 89.6 & &  \\
		\bottomrule
	\end{tabular}
	\caption{Performance of parsers using different transition systems on the Penn Treebank dataset. *:  Obtained from their published results online.\footnotemark} \label{tab:ptb}
\end{table}

%\cutout{We note that some system are trained with the dynamic oracle (dyn-oracle) \cite{goldberg2012dynamic}, a CRF-like loss (CRF-loss), or evaluated with beam search (beam size B). }
%For \cite{kiperwasser2016simple}, we include their best transition-based results which uses 11 BiLSTM vectors instead of 4 in our implementation, as well as their graph-based results.

%\todo{(exact results pending) some brief discussion about {\it arc-eager} variants, {\it arc-eager-S} does better than {\it arc-eager-R} probably because the former postpones transition decision to when more information is immediately available at the boundary of the stack and the buffer; but probably dyn-oracle is just the right way to go.}

\begin{table}[t]
\centering
\begin{tabular}{lcccc}
\toprule
	&	{\it aeS}	&	{\it asd}	&	{\it ah}	&	{\it aeR}\\
\midrule
{\it asw}	&	***/***	&	***/***	&	***/***	&	***/*** \\
{\it aeS}	&		&	-/-	&	-/-	&	*/- \\
{\it asd}	&		&		&	-/-	&	*/- \\
{\it ah}	&		&		&		&	-/- \\
\bottomrule
\end{tabular}
\caption{Significance test for transition systems. Each grid shows adjusted test result for UAS and LAS, respectively, showing whether the system on the row is significantly better than that on the column. ``***'' stands for $p<0.001$, ``**'' $p<0.01$, ``*'' $p<0.05$, and ``-'' $p \ge 0.05$.} \label{tab:sigtest}
\end{table}

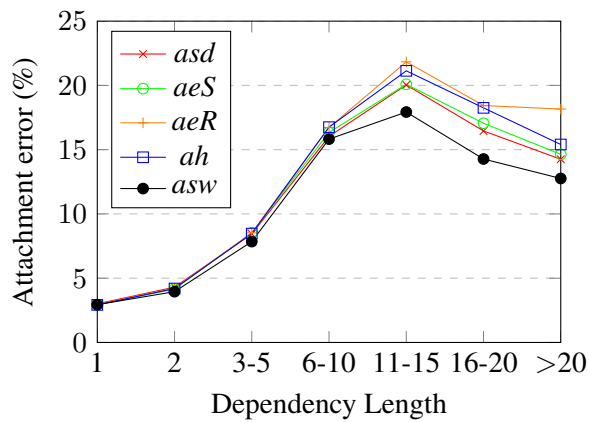
\begin{figure}
\begin{tikzpicture}
\begin{axis}[
xlabel={Dependency Length},
ylabel={Attachment error (\%)},
xmin=1, xmax=7,
ymin=0, ymax=25,
xtick={1,2,3,4,5,6,7},
ytick={0,5,10,15,20,25},
xticklabels={1,2,3-5,6-10,11-15,16-20,$>$20},
legend pos=north west,
ymajorgrids=true,
grid style=dashed,
]

\addplot[
color=red,
mark=x,
]
coordinates {
(1, 3.006240) (2, 4.301855) (3, 8.446207) (4, 16.030027) (5, 20.034930) (6, 16.457211) (7, 14.252874)
};
\addlegendentry{{\it asd}}

\addplot[
color=green,
mark=o,
]
coordinates {
(1, 2.933783) (2, 4.153828) (3, 8.523607) (4, 16.343596) (5, 20.084830) (6, 17.055655) (7, 14.636015)
};
\addlegendentry{{\it aeS}}

\addplot[
color=orange,
mark=+,
]
coordinates {
(1, 2.958921) (2, 4.220289) (3, 8.530057) (4, 16.723679) (5, 21.831337) (6, 18.432077) (7, 18.160920)
};
\addlegendentry{{\it aeR}}

\addplot[
color=blue,
mark=square,
]
coordinates {
(1, 2.921953) (2, 4.196121) (3, 8.472007) (4, 16.752185) (5, 21.132735) (6, 18.252543) (7, 15.402299)
};
\addlegendentry{{\it ah}}

\addplot[
color=black,
mark=*,
]
coordinates {
(1, 2.941176) (2, 3.951423) (3, 7.846362) (4, 15.820981) (5, 17.926647) (6, 14.272890) (7, 12.758621)
};
\addlegendentry{{\it asw}}

\end{axis}
\end{tikzpicture}
\caption{Parser attachment errors on PTB-SD binned by the length of the gold dependency.} \label{fig:length}
\end{figure}

For each pair of the systems we implemented, we studied the statistical significance of their difference by performing a paired test with 10,000 bootstrap samples on PTB-SD. The resulting p-values are analyzed with a 10-group Bonferroni-Holm test, with results shown in Table \ref{tab:sigtest}. We note that with almost the same implementation, \algname\ parsers significantly outperform those using traditional transition systems.
We also analyzed the performance of parsers on attachments of different distances. As shown in Figure \ref{fig:length}, \algname\ is equally accurate as existing systems for short dependencies, but is more robust for longer ones.

\footnotetext{\url{https://github.com/tensorflow/models/blob/master/syntaxnet/g3doc/universal.md}}

\diffadd{While \algname\ introduces direct long-distance transitions, it also shortens the overall sequence necessary to induce the same parse. A parser could potentially benefit from both factors: direct attachments could make an easier classification task, and shorter sequences limit the effect of error propagation. However, since the two effects are correlated in a transition system, precise attribution of the gain is out of the scope of this paper.}

%For English Universal Dependencies (EN-UD)w, we compare our models using the four transition systems with that of \cite{andor2016globally}.  .

\diffadd{\paragraph{Computational efficiency.} We study the computational efficiency of the \algname\ parser by comparing it to an \arceager\ parser. On the PTB-SD development set, the average transition sequence length per sentence of \algname\ is 77.5\% of that of \arceager. At each step of parsing, \algname\ needs to evaluate only about 1.24 times the number of transition candidates as \arceager, which results in very similar runtime. In contrast, beam search with beam size 2 for \arceager\ requires evaluating 4 times the number of transition candidates compared to greedy parsing, which results in a UAS 0.14\% worse and LAS 0.22\% worse for \arceager\ compared to greedily decoded \algname.}

\subsection{Linguistic Analysis}

%\begin{table}[t]
%	\centering
%\begin{tabular}{lrrr}
%	\toprule
%	& {\it asw}  & {\it aeS} & {\it asd}\\
%	\midrule
%	PP attachment & 260 & 272 & 281 \\
%	Conjunction attachment & 131 & 150 & 136 \\
%	Adverbial attachment &  101 & 97 & 106 \\
%	\bottomrule
%\end{tabular}
%\caption{Common parser errors on EN-UD.} \label{tab:errors_ud}
%\end{table}

We automatically extracted all labelled attachment errors by error type (incorrect attachment or relation), and categorized a few top parser errors by hand into linguistic constructions.
Results on PTB-SD are shown in Table \ref{tab:errors}.\footnote{We notice that for some examples the parsers predicted a {\it ccomp} (complement clause) attachment to verbs ``says'' and ``said'', where the CoreNLP output simply labelled the relation as {\it dep} (unspecified). For other examples the relation between the prepositions in ``out of'' is labelled as {\it prep} (preposition) instead of {\it pcomp} (prepositional complement). We suspect this is due to the converter's inability to handle certain corner cases, but further study is warranted.} We note that the \algname\ parser improves accuracy on prepositional phrase (PP) and conjunction attachments, while it remains comparable to other parsers on other common errors. Analysis on EN-UD shows a similar trend. As shown in the table, there are still many parser errors unaccounted for in our analysis. We leave this to future work.
%The two left-to-right transition systems also suffer less from errors in prepositional phrase attachments, compared to the bottom-up \arcstandard.
\clearpage
\begin{table}
	\centering
	\begin{tabular}{@{}lrrr@{}}
		\toprule
		& {\it asw}  & {\it aeS} & {\it asd}\\
		\midrule
		PP attachment & 545 & 569 & 571 \\
		Noun/Adjective confusion & 221 & 230 & 221\\
		Conjunction attachment & 156 & 170 & 164 \\
		Adverbial attachment &  123 & 122 & 143 \\
		\midrule
		Total Errors & 3884 & 4100 & 4106 \\
		%Conjunct attachment & {\bf 38} & 47 & 40 \\
		\bottomrule
	\end{tabular}
	\caption{Common parser errors on PTB-SD. \diffadd{The top 4 common errors are categorized and shown in this table. Errors not shown are in a long-tail distribution and warrant  analyses in future work.}} \label{tab:errors}
\end{table}

%\todo{(low-priority, also results pending) analysis of effect of beam size}
\section{Related Work} \label{sec:related_work}

\diffsub{The most similar previous work is \cite{attardi2006experiments}. A notable difference is that \cite{attardi2006experiments} is designed for non-projective parsing, and retains tokens between the head and dependent after inducing a long arc. Moreover, limited by the sparse feature-based classifier used, direct attachments are only made with up to the top 3 tokens in the stack.}

\diffadd{Previous work has also explored augmenting transition systems to facilitate longer-range attachments. \citet{attardi2006experiments} extended the \arcstandard\ system for non-projective parsing, with arc-inducing transitions that are very similar to those in \algname. A notable difference is that their transitions retain tokens between the head and dependent. \citet{fernandez2012improving} augmented the \arceager\ system with transitions that operate on the buffer, which shorten the transition sequence by reducing the number of Shift transitions needed. However, limited by the sparse feature-based classifiers used, both of these parsers just mentioned only allow direct attachments of distance up to 3 and 2, respectively. More recently, \citet{sartorio2013transition} extended \arcstandard\ with transitions that directly attach to left and right ``spines'' of the top two nodes in the stack. While this work shares very similar motivations as \algname, it requires additional data structures to keep track of the left and right spines of nodes.  This transition system also introduces \emph{spurious ambiguity} where multiple transition sequences could lead to the same correct parse, which necessitates easy-first training to achieve a more noticeable improvement over \arcstandard. In contrast, \algname\ can be easily implemented given the parser state alone, and does not give rise to spurious ambiguity.}

For a comprehensive study of transition systems for dependency parsing, we refer the reader to \cite{bohnet2016generalized}, which proposed a generalized framework that could derive all of the traditional transition systems we described by configuring the size of the active token set \diffsub{$K$ }and the maximum arc length\diffsub{ $D$}, among other control parameters. \diffsub{\algname\ can be considered a system where $K \in \{2,\ldots, n+1\}$, and $D=n$ under that framework, where $n$ is the total number of tokens (except $\langle$root$\rangle$) in that framework.} \diffadd{However, this framework does not cover \algname\ in its original form, as the authors limit each of their transitions to reduce at most one token from the active token set (the buffer). On the other hand, the framework presented in \cite{gomez2013divisible} does not explicitly make this constraint, and therefore generalizes to \algname. However, we note that \algname\ still falls out of the scope of existing discussions in that work, by introducing multiple Reduces in a single transition.}

\section{Conclusion}\label{sec:conclusion}

%\todo{This can probably be further shortened}

In this paper, we introduced \algname, a novel transition system for dependency parsing. We also performed linguistic analyses on parser outputs and showed \algname\ parsers reduce errors in conjunction and adverbial attachments compared to parsers using traditional transition systems. 
%We showed that with the same neural network implementation of the parser, parsers using \algname outperform those using previous transition systems by reducing parsing errors, especially in categories such as \todo{PP attachment, CC attachment, etc.} We also compared empirically the effectiveness of the three traditional transition systems, {\it arc-standard}, {\it arc-eager}, and {\it arc-hybrid}, which is the first side-by-side direct comparison to the best of our knowledge. \todo{Well... Really just my knowledge. Chris? Danqi?}

\section*{Acknowledgments}
We thank Timothy Dozat, Arun Chaganty, Danqi Chen, and the anonymous reviewers for helpful discussions. Stanford University gratefully acknowledges the support of the Defense Advanced Research Projects Agency (DARPA) Deep Exploration and Filtering of Text (DEFT) Program under Air Force Research Laboratory (AFRL) contract No.\ FA8750-13-2-0040. Any opinions, findings, and conclusion or recommendations expressed in this material are those of the authors and do not necessarily reflect the view of the DARPA, AFRL, or the US government.

%\bibliography{acl2017}
\bibliography{acl2017}
\bibliographystyle{acl_natbib}

\appendix
\section{Model and Training Details}\label{sec:model_training}

\begin{figure}
	\centering
	\includegraphics[width=.475\textwidth]{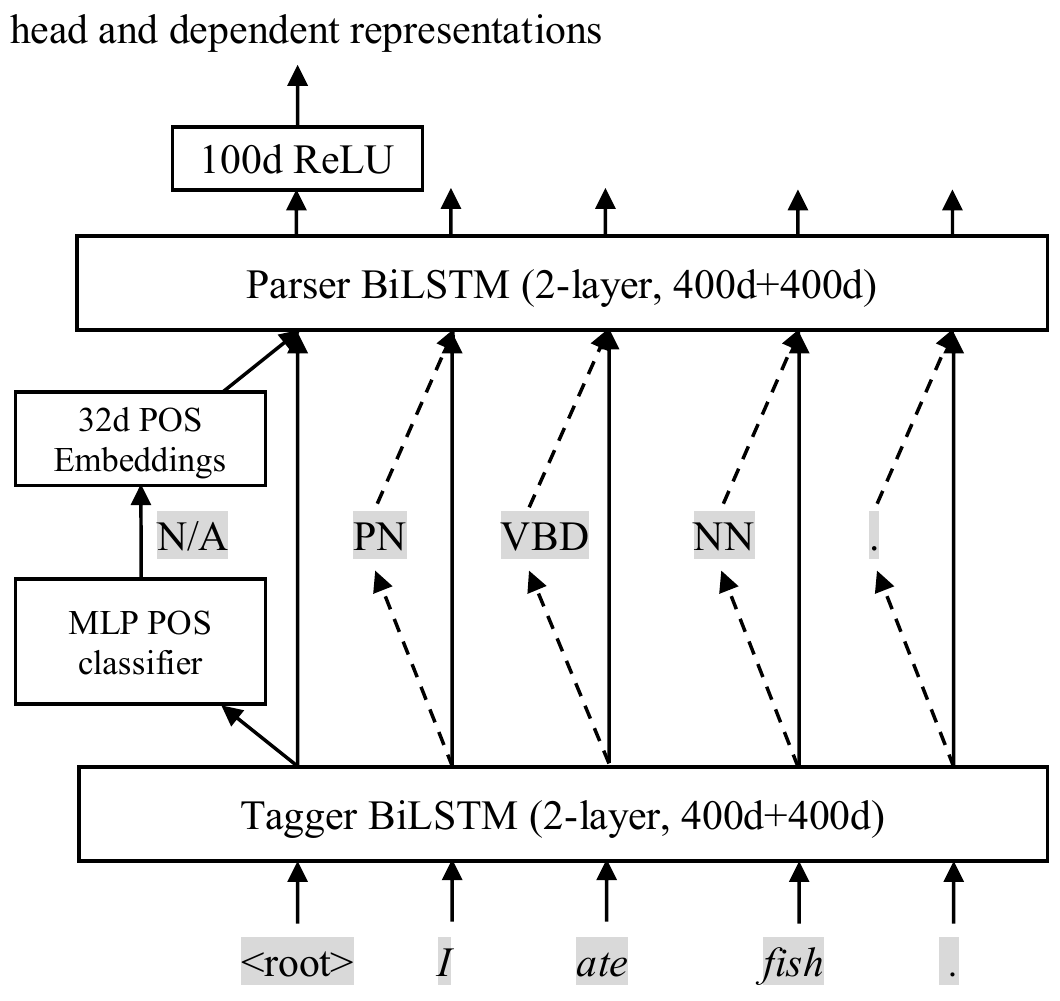}
	\caption{Illustration of the model.} \label{fig:model}
\end{figure}

Our model setup is similar to that of \citep{kiperwasser2016simple} (See Figure \ref{fig:model}). We employ two blocks of bidirectional long short-term memory (BiLSTM) networks \cite{hochreiter1997long} that share very similar structures, one for part-of-speech (POS) tagging, the other for parsing. Both BiLSTMs have 400 hidden units in each direction, and the output of both are concatenated and fed into a \diffsub{100-dimensional} dense layer of rectified linear units (ReLU) before 32-dimensional representations are derived as classification features. As the input to the tagger BiLSTM, we represent words with 100-dimensional word embeddings, initialized with GloVe vectors \cite{pennington2014glove}.\footnote{We also kept the vectors of the top 400k words trained on Wikipedia and English Gigaword for a broader coverage of unseen words.} The output distribution of the tagger classifier is 
%and pass the word representations through a 2-layer bidirectional long short-term memory (BiLSTM) network \citep{hochreiter1997long}. The output of the two directions are concatenated, and fed into a classifier to predict part-of-speech (POS) tags for each word. The output distribution of the classifier is 
used to compute a weighted sum of 32-dimensional POS embeddings, which is then concatenated with the output of the tagger BiLSTM (800-dimensional per token) as the input to the parser BiLSTM. For the parser BiLSTM, we use two separate sets of dense layers to derive a ``head'' and a ``dependent'' representation for each token.
%The output of this second BiLSTM is concatenated and passed through a fully connected layer with rectified linear unit activations, to form a 32-dimensional ``head'' representation and a 32-dimensional ``dependent'' representation for each word. 
These representations are later merged according to the parser state to make transition predictions. 

For traditional transition systems, we follow \cite{kiperwasser2016simple} by featurizing the top 3 tokens on the stack and the leftmost token in the buffer. To derive features for each token, we take its head representation $v_{\text{head}}$ and dependent representation $v_{\text{dep}}$, and perform the following biaffine combination
\begin{align}
v_{\text{feat}, i} &= [f(v_{\text{head}}, v_{\text{dep}})]_i \nonumber \\
\begin{split}
&= \text{ReLU}\bigl(v_{\text{head}}^\top W_i v_{\text{dep}} + b_i^\top v_{\text{head}}\\
   &\phantom{xxx\text{ReLU(}} + c_i^\top v_{\text{dep}} + d_i\bigr)， \label{eqn:biaffine}
\end{split}
\end{align}
where $W_i\in \mathbb{R}^{32\times 32}$, $b_i, c_i\in \mathbb{R}^{32}$, and $d_i$ is a scalar for $i=1, \ldots, 32$. The resulting 32-dimensional features are concatenated as the input to a fixed-dimensional softmax classifier for transition decisions. %The parsers are trained to maximize the log likelihood of the desired transition sequence.

For \algname, we featurize for each arc-inducing transition with the same composition function in Equation (\ref{eqn:biaffine}) with $v_{\text{head}}$ of the head token and $v_{\text{dep}}$ of the dependent token of the arc to be induced. %. For each arc-inducing transition, 
\diffsub{For each transition, the feature is used to generate $|L|$ scores through an affine transform, where $L$ is the set of all possible arc labels. }
For Shift, we simply combine $v_{\text{head}}$ and $v_{\text{dep}}$ of the leftmost token in the buffer with the biaffine combination, and obtain its score 
%For Shift, the score is obtained 
by computing the inner-product of the feature and a vector. At each step, the scores of all feasible transitions are normalized to a probability distribution by a softmax function.

In all of our experiments, the parsers are trained to maximize the log likelihood of the desired transition sequence, along with the tagger being trained to maximize the log likelihood of the correct POS tag for each token.

To train the parsers, we use the {\sc Adam} optimizer \citep{kingma2014adam}, with $\beta_2 = 0.9$, an initial learning rate of 0.001, and minibatches of size 32 sentences. Parsers are trained for 10 passes through the dataset on PTB-SD. We also find that annealing the learning rate by a factor of 0.5 for every pass after the 5th helped improve performance. For EN-UD, we train for 30 passes, and anneal the learning rate for every 3 passes after the 15th due to the smaller size of the dataset. For all of the biaffine combination layers and dense layers, we dropout their units with a small probability of 5\%. Also during training time, we randomly replace \diffadd{10\% of the input} words \diffsub{in the training set }by an artificial $\langle$UNK$\rangle$ token, which is then used to replace all unseen words in the development and test sets. Finally, we repeat each experiment with 3 independent random initializations, and use the average result for reporting and statistical significance tests.

The code for the full specification of our models and aforementioned training details \diffadd{are available at \gitrepo}.

\end{document}